\documentclass{article}
\usepackage{ijcai19}

\usepackage{times}
\usepackage{soul}
\usepackage{url}
\usepackage[hidelinks]{hyperref}
\usepackage[utf8]{inputenc}
\usepackage[small]{caption}
\usepackage{graphicx}
\usepackage{amsmath}
\usepackage{booktabs}
\usepackage{algorithm}
\usepackage{algorithmic}
\usepackage{acronym}
\usepackage[skip=0pt]{caption}
\usepackage{amsfonts}
\usepackage{color}
\usepackage{natbib}
\usepackage{fancyhdr}
\usepackage{hhline}

\acrodef{BBC}{Background Based Conversation}
\acrodef{RC}{Reading Comprehension}
\acrodef{BiDAF}{Bi-directional attention flow}
\acrodef{GTTP}{Get To The Point}
\acrodef{HRED}{Hierarchical Recurrent Encoder-decoder Architecture}
\acrodef{S2S}{Sequence to Sequence}
\acrodef{S2SA}{Sequence to Sequence with Attention}
\acrodef{GRU}{Gated Recurrent Units}
\acrodef{b2c}{background-to-context}
\acrodef{c2b}{context-to-background}
\acrodef{LSTM}{Long Short-Term Memory}
\acrodef{RNN}{Recurrent Neural Network}
\acrodef{Ptr-Net}{Pointer Network}
\acrodef{MMI}{Maximum Mutual Information}
\acrodef{CaKe}{Context-aware Knowledge Pre-selection}

\DeclareMathOperator*{\softmax}{softmax}

\title{Improving Background Based Conversation with \\ Context-aware Knowledge Pre-selection}

\author{
Yangjun Zhang
\and
Pengjie Ren\And
Maarten de Rijke
\affiliations
University of Amsterdam
\emails{
\{y.zhang6,
p.ren,
derijke\}@uva.nl}
}

\usepackage{multirow}
\begin{document}

\maketitle

\begin{abstract}
\acp{BBC} have been developed to make dialogue systems generate more informative and natural responses by leveraging background knowledge. 
Existing methods for \acp{BBC} can be grouped into two categories: \emph{extraction-based} methods and \emph{generation-based} methods.
The former extract spans from background material as responses that are not necessarily natural.
The latter generate responses that are natural but not necessarily effective in leveraging background knowledge.
In this paper, we focus on \emph{generation-based} methods and propose a model, namely \ac{CaKe}, which introduces a pre-selection process that uses dynamic bi-directional attention to improve knowledge selection by using the utterance history context as prior information to select the most relevant background material. 
Experimental results show that our model is superior to current state-of-the-art baselines, indicating that it benefits from the pre-selection process, thus improving informativeness and fluency.
\end{abstract}


\section{Introduction}
Dialogue systems have attracted great attention recently. Many methods have achieved promising results \citep{sutskever2014sequence, bahdanau2014neural}. 
However, many challenges, including lack of diversity, limited informativeness and inconsistency, still remain.
Among them, the lack of informativeness may well be the most important one. 
The responses of a chat bot may be diverse without being informative (e.g., \emph{I can't tell you}, \emph{I'm not sure}, \emph{I don't have a clue}).
\acf{BBC} is a method that utilizes external knowledge to solve this problem, thus decreasing the number of bland and deflective responses~\citep{moghe2018towards}. 
Given background material and a conversation, the \acf{BBC} task is to generate responses by referring to background information and considering the dialogue history context at the same time. 
The background material may be semi-structured information or free text; in this paper, background material is provided in the form of free text. 
\acp{BBC} have demonstrated a potential for generating more informative responses. 

There are two main approaches to \acp{BBC}, extraction-based methods and generation-based methods. 
Extraction-based methods extract text segments from the background as responses. 
Hence, the responses are informative.
Generation-based methods generate sequences based on a encoder-decoder mechanism and the responses are natural and fluent. 
For extraction-based methods, models including the \ac{BiDAF} model \citep{seo2016bidirectional} have been proposed for selecting the best matching positions of the tokens from the target context.
However, responses produced by extraction-based methods are directly copied from background sentences.
As a result, they are neither fluent nor natural. 
Generation-based methods include \ac{HRED} \citep{serban2016building} and \ac{GTTP} \citep{see2017get, moghe2018towards}; they are not always effective at leveraging background knowledge and may return responses with inappropriate background knowledge. 

\begin{figure}[ht]
  \centering
  \includegraphics[width=\linewidth]{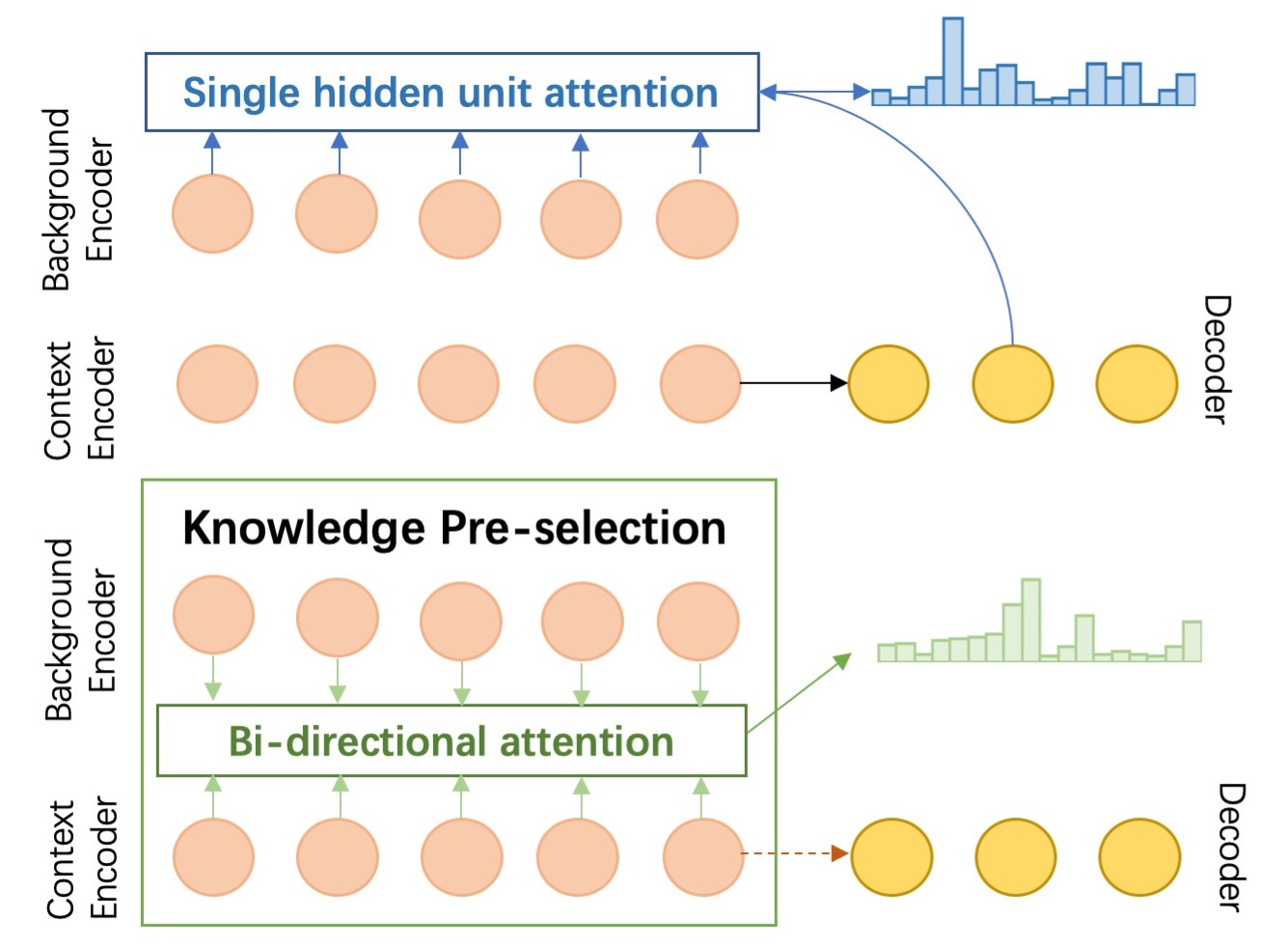}
  \caption{Knowledge pre-selection.}
  \label{fig:intro}
\end{figure}

The motivation behind our model is to improve the process of utilizing background knowledge. 
The \ac{GTTP} model \citep{see2017get, moghe2018towards} selects background knowledge by using a hidden state at each decoding time step as a query to select background knowledge. 
For each query, it is supposed to comprise features of the corresponding response token, features from the previous response tokens, as well as features passed on from the utterance history context. 
However, the query may not contain all information from the utterance history since the information attenuated rapidly \citep{cho2014properties} during state transfer especially when the length of the input utterance history context of the encoder increases, as the red line demonstrates in Figure~\ref{fig:intro}. 



The utterance history contains rich information about the conversation and it is also highly relevant to the response. 
Therefore, we propose a neural model, namely \acfi{CaKe}, which introduces a knowledge pre-selection step to offset the problem of the current state-of-art model.
\ac{CaKe} utilizes utterance history context as query directly to facilitate better selection of knowledge. 
There are two steps to implement knowledge pre-selection in \ac{CaKe}. 
To start, we choose the encoder state of the utterance history context as a query to select the most relevant knowledge. 
Then, we employ a modified version of \ac{BiDAF} that combines background-to-context attention and context-to-background attention to point out the most relevant token positions of the background sequence to cope with long background text.

The main contribution of our model is to propose a pre-selection module. 
The pre-selection process leverages the utterance history of the corresponding response as a query to search a set of positions of the background knowledge sentence and find the most relevant knowledge. 
Meanwhile, we use a modified bi-attention flow mechanism to achieve the pre-selection process. 
Experiments show that \ac{CaKe} significantly outperforms the strong generation-based baseline, \ac{GTTP}.
Although \ac{CaKe} cannot beat extraction-based baseline, \ac{BiDAF}, to a large margin, \ac{CaKe} can generate more natural responses than \ac{BiDAF}.


\section{Related Work}
\subsection{Extraction-based Methods}
Extraction-based methods to \ac{BBC} are originally derived from \ac{RC} tasks~\citep{rajpurkar2016squad}, where the answer could be picked from a set of token positions of the input sequence.
\citet{vinyals2015pointer} propose a \ac{Ptr-Net} model that uses attention as a pointer to select a token in an input sequence so as to generate an output sequence where some tokens come from the input sequence. 
\citet{wang2016machine} extend this work by combining match-\ac{LSTM} and \ac{Ptr-Net}. 
\citet{seo2016bidirectional} introduce \ac{BiDAF} to improve the extraction of the context span by pointing to the start point and the end point of the relevant span. 
\citet{lee2016learning} and \citet{yu2016end} predict answers by ranking text spans, where the token positions of the span are continuous, within background passages. 
\citet{wang2017gated} predict the boundary of the answer span by a self-matching mechanism.
Extraction-based methods are better at locating the right background span than generation-based methods \citep{moghe2018towards}. 
Nevertheless, extraction-based methods are not suitable for \acp{BBC} as \acp{BBC} do not have standard answers like those in \ac{RC} tasks and responses based on fixed extraction of the background are not natural or fluent enough for conversation tasks.  

\subsection{Generation-based Methods}
\ac{S2S} learning methods \citep{sutskever2014sequence}, later extended to \ac{S2SA}~\citep{bahdanau2014neural}, are the basis of most generation-based methods.
Although generation-based methods achieve good results on different conversation tasks, many challenges remain.

First, response diversity is to be improved. 
Lots of methods have been proposed to solve this issue. 
To increasediversity, \citet{li2015diversity} present \ac{MMI} as the objective function in neural models to decrease generic response sequences and increase varied and interesting outputs. 
\citet{serban2016building} propose \ac{HRED}, which uses a two-level hierarchy, including word level and dialogue turn level, to exploit long-term text. 
Later, \citet{serban2017hierarchical} add a high-dimensional stochastic latent variable to the target to extend \ac{HRED}, aiming to generate responses with more diversified content and reduce blandness. 
\citet{zhang2018generating} explicitly optimize a variational lower bound on pairwise mutual information between query and response to boost diversity during training.

Second, response informativeness is an important issue. 
Many methods adding background knowledge and common sense to conversations have been proposed for mitigating blandness. 
Most of the existing conversational datasets are not labelled with relevant knowledge, so it is difficult to apply large datasets to the model training. 
As a result, most models need to do knowledge selection before training the model, such as knowledge diffusion \citep{liu2018knowledge} and graph attention \citep{zhou2018commonsense}. 
These methods use knowledge datasets that are separate from the conversations.
Recently, several datasets have become available where conversations are generated based on background knowledge.
\citet{moghe2018towards} build a dataset for \acp{BBC} and conduct experiments with several methods. 
The datasets of Persona-chat \citep{zhang2018personalizing} and Wizard-of-Wikipedia \citep{dinan2018wizard} are similar to \citet{moghe2018towards}.

Although the background knowledge available in \acp{BBC} has a low degree of redundancy, selecting the most relevant background is important to improve informativeness.
\citet{moghe2018towards} employ \ac{GTTP} proposed by \cite{see2017get} to copy tokens from background knowledge at each generation timestamp.
\citet{lian2019learning} use posterior knowledge distribution to guide knowledge selection. 
The selection of background material plays a vital role in generating informative responses. 
However, the crucial role of context history in selecting appropriate background has not been fully explored by current methods. 
Unlike previously proposed methods, in order to encourage informative and non-deflective responses, our proposed model leverages the context as prior context to do pre-selection of the background knowledge.

%

\begin{figure*}
  \centering
  \includegraphics[width=16cm]{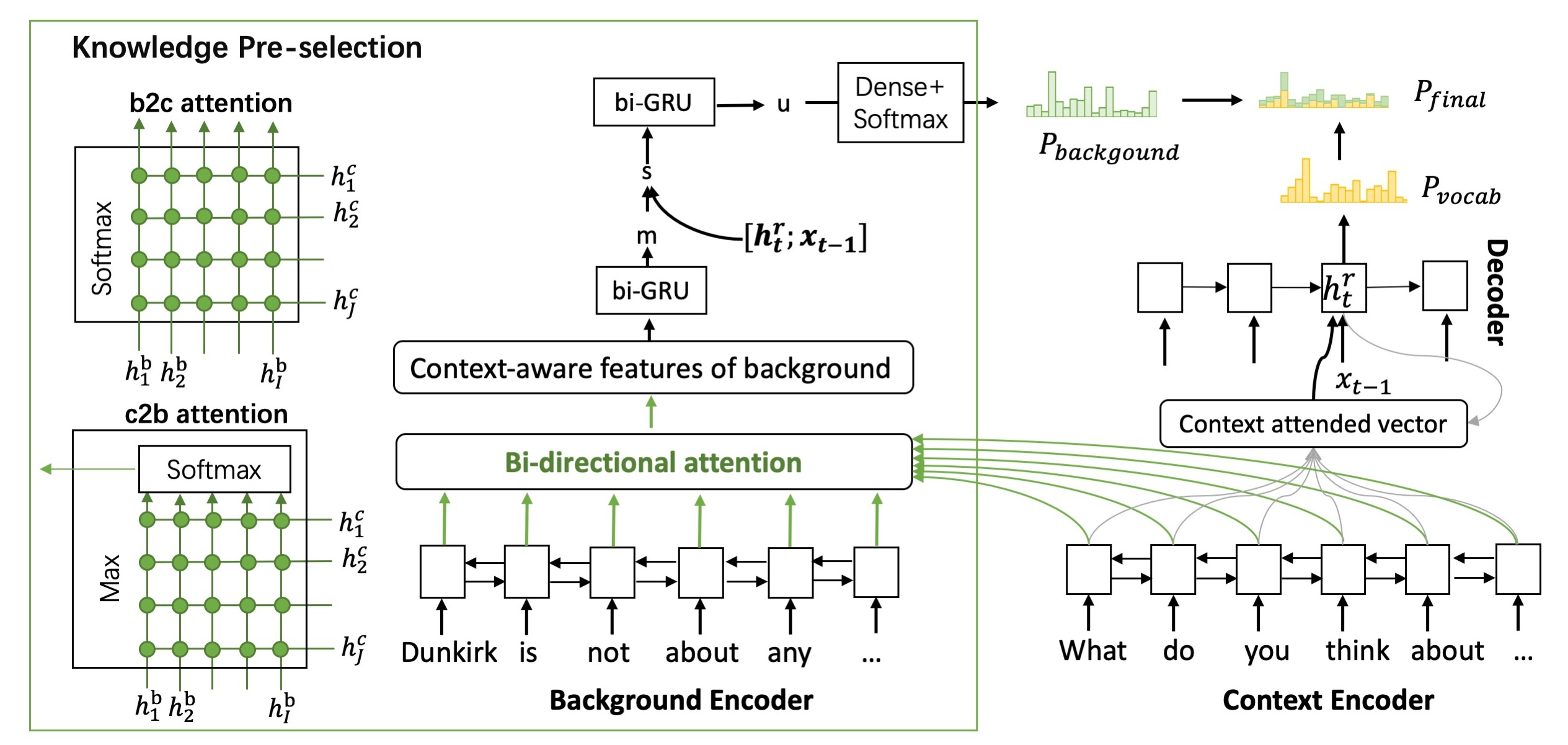}
  \caption{Overview of our model, \acf{CaKe}.}
  \label{fig:method}
\end{figure*}

\section{Methodology}
Given a background in the form of free text and the current utterance history context, \ac{BBC} aims to generate an utterance as the next response.
Formally, let $b=(b_1,b_2,\ldots,b_i\ldots,b_I)$ represent the words in the background knowledge, a current utterance history context in the form of $c=(c_1,c_2,\ldots,c_j,\ldots,c_J)$, and the task of \ac{BBC} is to generate response $r=(x_1,x_2,\ldots,x_t,\ldots,x_T)$ based on $b$ and $c$. 

In this section we introduce our model \ac{CaKe} for \ac{BBC}. 
The overview of \ac{CaKe} is shown in Figure \ref{fig:method}.
First, we use two encoders to encode background and context. 
Second, for knowledge pre-selection, we select the background related to the context. 
To achieve this, we choose the encoder state of the context as query to select the most relevant knowledge from the background knowledge. 
We use a modified version of \ac{BiDAF} \citep{seo2016bidirectional} which combines background-to-context attention and context-to-background attention to point out the most relevant token position of the background sequence. 
The pre-selection module forms context-aware background distribution.
Third, for the generation part, the generator generates vocabulary distribution with global attention \citep{bahdanau2014neural}. 
Lastly, We combine this pre-selector part with the generator to generate the final output of each decoding time step. 
We get the response based on the probability of generating token from the vocabulary or copying token from the background.

\subsection{Background and Context Encoder}
The word embeddings are used to map words to high-dimensional vector space. 
We apply random initialized word embeddings.

The background and context encoders encode background and context embeddings into $h^b=(h^b_1,h^b_2,\ldots,h^b_i,\ldots,h^b_I)$ and $h^c=(h^c_1,h^c_2,\ldots,h^c_j,\ldots,h^c_J)$ respectively. 
We use bidirectional \acp{RNN} and concatenate the outputs of two \acp{RNN} for the two encoders. 
Therefore, we get $h^b \in \mathbb{R}^{2d \times I}$ for the background and $h^c \in \mathbb{R}^{2d \times J}$ for the context.

\subsection{Knowledge Pre-selection} 
The selector is used for pre-selecting background words and form context-aware background distribution.
We combine background-to-context and context-to-background attention. The original \ac{BiDAF} \citep{seo2016bidirectional} trains start and end span position. \citet{moghe2018towards} uses this model to find the model relevant span from the background knowledge by two positions, while we only use the start span position as the most relevant token index. 
The similarity score between context and background is calculated by: 
    \begin{equation}
        \begin{split}
          score_{ij}=S(h^b_{:i},h^c_{:j}),
        \end{split}
    \end{equation}
    where $h^b_{:i}$ is the $i^{\text{\tiny th}}$ column vector of $h^b$ and $h^c_{:j}$ is the $j^{\text{\tiny th}}$ column vector of $h^c$. 
    Meanwhile, $S$ is defined as:
    \begin{equation}
        \begin{split}
          S(h^b,h^c)=w^T[h^b;h^c;h^b \odot h^c],
        \end{split}
    \end{equation}
     where $w^T$ is a trainable weight vector. 
     Then, we feed $score_{ij}$ to softmax function to get the attention and the corresponding vector.
    
First, \ac{b2c} attention reflects which context words are most relevant to each background word. 
$\alpha_{i}$ represents the attention weights on the context words by the $i$-th background word, 
    where $\sum_{i}a_{ij}=1$. 
    The attention vector of context is computed by
    \begin{equation}
        \begin{split}
          \widetilde{h}^c_{:i}=\sum\limits_j \alpha_{ij}h^c_{:j},
        \end{split}
    \end{equation}
    where $\alpha_{i}$ is computed by normalizing $S_{i:}$ by a $\softmax$ function. $\widetilde{h}^c \in \mathbb{R}^{2d \times I}$.
    
    Second, \ac{c2b} attention computes which background words are most relevant to each context word. 
    The attention weights on the background words are calculated by $\beta=\softmax(\max_{col}(S)$.
    Then the attended background vector is computed by:
    \begin{equation}
           \widetilde{h}^b_{:i}=\sum\limits_i \beta_{i}h^b_{:i},
    \end{equation}
    where $\widetilde{h}^b \in \mathbb{R}^{2d \times I}$.
    
    Finally, contextual embeddings and the attention vectors are combined to yield $g$, where $g$ is defined by:
    \begin{equation}
            g_{:i}=\eta(h^b_{:i},\widetilde{h}^c_{:i},\widetilde{h}^b_{:i}),
    \end{equation}
    where $\eta$ is a trainable vector. 
    We use simple concatenation in our experiments: $\eta(h^b,h^c,\widetilde{h}^b)=[h^b,\widetilde{h}^c,h \odot \widetilde{h}^c,h^b \odot \widetilde{h}^b]$.
    $g$ is the static context-aware background representations. We use a bi-\ac{RNN} layer for $g$, and get $m$, which captures the interaction among the background words conditioned on the context. 
    Then we concatenate $m$ with $h^r_t$ and $x_{t-1}$ to generate $s$. 
    $s$ is fed into bi-\ac{RNN} to generate $u$.
    Finally, background distribution is calculated by: 
     \begin{equation}
            P_{background}=\softmax(w^T_{p1}[g;m;s;u]),
    \end{equation}
    where $W^T_{p1}$ is a trainable factor.

    \subsection{Generator}
    For the generator, we generate the response token, which is the vocabulary distribution, with attention. 
    For the generator module, the hidden state of response is: $h^r=(h^r_1$, $h^r_2$, \ldots, $h^r_t$,\ldots, $h^r_T)$, where $h^r_t$ is the state of the decoder at the current time step. 
    The final decoder hidden state aware representation of the context is the attention weighted sum of context $c_t$. 
    The representation of context is $h^c_1$, $h^c_2$, \ldots, $h^c_j$, \ldots, $h^c_J$, where $J$ is the total length of the context.  The final attention weighted sum of the context is calculated as follows:
     \begin{equation}
        \begin{split}
        e^t_i& =v^T\tanh(W_ch^c_j+Vh^r_t+b_c)\\
        \gamma^t&=\softmax(e^t)\\
        c_t&=\sum_j \gamma^t_j h^c_j,
        \end{split}
    \end{equation}
    where $h^r_t$ is the current state of the decoder.
    
    The generator then uses $c_t$, $s_t$ and $x_t$ to generate $P_{vocab}$, and the generation probability $p_{gen}$ is calculated as follows:
      \begin{equation}
            P_{gen}=\sigma(w^T_cc_t+w^T_ss_t+w^T_xx_t+b_{gen}),
    \end{equation}
    where $w^T_c, w^T_s, w^T_x, b_{gen}$ are trainable parameters. 
    $p_{gen}$ is used as a switch to choose between generating a word by sampling from the vocabulary distribution $P_{vocab}$ or copying a word from the background by sampling from background distribution $P_{background}$.
    
    \subsection{Mixture and Loss}
    
    The mixture is used to mix the results of knowledge pre-selection and generation. We obtain the following final distribution over the extended vocabulary:
        \begin{equation}
          P_{final}(w)=p_{gen}P_{vocab}(w)+(1-p_{gen})P_{background}.
    \end{equation}
    Let $P(w)=P_{final}(w)$. During training, the training loss of time step $t$ is the negative log likelihood of the target word $w^*_t$ for that time step:
     \begin{equation}
           loss_t=-\log P(w^*_t).
    \end{equation}
     The training loss of the whole sequence $loss$ and the whole datasets $L(\theta)$ are:
    \begin{equation}
        \begin{split}
          loss&=\frac{1}{T}\sum^T_{t=0}loss_t\\
          L(\theta)&=\sum^N_{n=0}loss \label{eq:loss},
        \end{split}
    \end{equation}
   where $\theta$ is the set of all trainable weights, $N$ is the number of total samples in the dataset. 

\section{Experiment Setup}

\subsection{Datasets}
Recently, several datasets have been released for \acp{BBC} \citep{zhang2018personalizing, moghe2018towards, dinan2018wizard}. 
We choose the Holl-E dataset released by \citet{moghe2018towards}. 
The data contains background documents of 921 movies and 9071 conversations.
The background documents of the movies contain four parts: review, plot, comment, meta-data or fact table. 
The conversations have two speakers. 
For the first speaker, the background documents are not available while the second one could use knowledge from background documents during chatting.
We use two versions of background documents: oracle background and 256-word background. 
Oracle background uses the actual resource part from the background documents. 
The 256 words background is generated by truncating the background sentences.

\begin{table*}[ht]
  \caption{Automatic evaluation results.}
  \label{tab:result}
  \centering
  \begin{tabular}{lcccccccc}
    \toprule
    \multirow{2}{4em}{\bf Methods}
     & \multicolumn{2}{c}{\bf BLEU} & \multicolumn{2}{c}{\bf ROUGE-1} &\multicolumn{2}{c}{\bf ROUGE-2} & \multicolumn{2}{c}{\bf ROUGE-l}  \\
     \cmidrule(r){2-3} \cmidrule(r){4-5}\cmidrule(r){6-7}\cmidrule(r){8-9}
          & \bf SR & \bf MR & \bf SR & \bf MR & \bf SR & \bf MR & \bf SR & \bf MR \\\midrule
    \multicolumn{9}{c}{\emph{No background}}\\\midrule
    {S2S}  & 4.63 & 7.01 & 26.91 & 30.50 & 9.34 & 11.36 & 21.58 & 24.99 \\
    {HRED}  & 5.23 & 5.38 & 24.55 & 25.38 & 7.66 & 8.35 & 18.87 & 19.67 \\\midrule
    \multicolumn{9}{c}{\emph{256 words background}}\\\midrule
    {S2SA}  & 11.71 & 12.76 & 26.36 & 30.76 & 13.36 & 16.69 & 21.96 & 25.99 \\
    {BiDAF}  & \textbf{27.44} & \textbf{33.40} & 38.79 & 43.92 & \textbf{32.91} & \textbf{37.86} & 35.09 & 40.12 \\
    {GTTP}  & 13.97 & 18.63 & 29.82 & 35.02 & 17.98 & 22.54 & 25.14 & 33.01 \\
    {\ac{CaKe}}  & 26.17 & 29.49 & \textbf{41.26} & \textbf{45.81} & 29.43 & 34.00 & \textbf{36.01} & \textbf{40.79} \\\midrule
    \multicolumn{9}{c}{\emph{Oracle background}}\\\midrule
    {S2SA}  & 12.26 & 13.11 & 27.51 & 31.89 & 13.98 & 17.55 & 22.85 & 27.03 \\
    {BiDAF}  & 24.93 & \textbf{32.21} & 35.60 & 42.40 & 29.48 & \textbf{36.54} & 31.72 & 38.39  \\
    {GTTP}  & 15.32 & 17.32 & 30.60 & 35.78 & 17.18 & 21.89 & 24.99 & 29.77 \\
    {\ac{CaKe}} & \textbf{26.02} & 31.16 & \textbf{42.82} & \textbf{48.65} & \textbf{30.37} & \textbf{36.54} & \textbf{37.48} & \textbf{43.21} \\
    \bottomrule
  \end{tabular}
\end{table*}

\begin{figure*}[!htb]
  \centering
  \includegraphics[width=17.5cm]{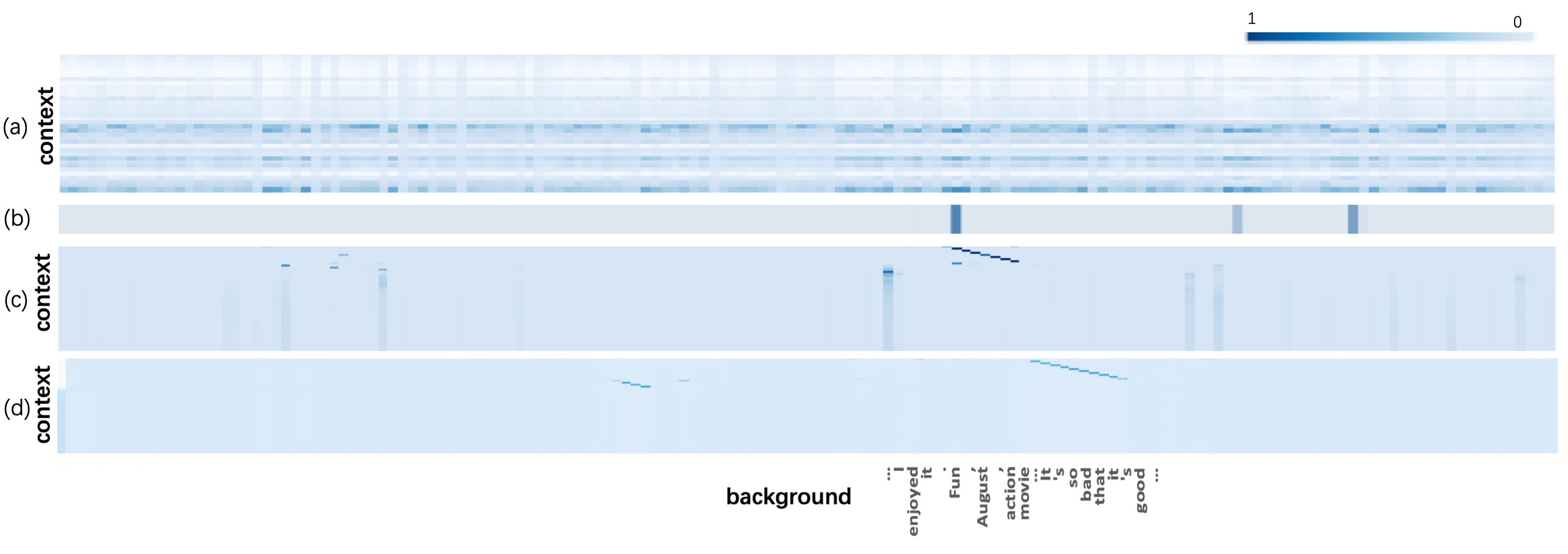}
  \caption{Knowledge selection visualization. (a) Background-to-context attention of CaKe; (b) Context-to-background attention of CaKe; (c) final pre-selection distribution on the background of CaKe; (d) knowledge distribution of GTTP.}
  \label{fig:attention}
\end{figure*}

\subsection{Implementation Details}
We use \ac{GRU} \citep{chung2014empirical} as the \ac{RNN} cell.
The dimension of word embedding is 128 according to a rule of thumb and the \ac{GRU} cell has a 256-dimensional hidden size. 
We use a vocabulary of 45k words.
We limit the context length of all models to 120. 
We train all the models for 30 epochs and the loss converges. 
The best model is selected based on the BLEU and ROUGE metric. 
We use gradient clippings with a maximum gradient norm of 2 and do not use any form of regularization. 
The word embeddings are learned from scratch during training. 
We use Adam optimizer with batch size 32 and learning rate 0.001. 
\ac{CaKe} is written in PyTorch and trained on four GeForce GTX TitanX GPUs.

\subsection{Baselines and Evaluation Metrics}
We compare \ac{CaKe} with several state-of-the-art methods.
\textbf{S2S} \citep{sutskever2014sequence} generates response from context with a simple sequence-to-sequence structure. 
\textbf{HRED} \citep{serban2016building} is a model using two-level hierarchy to encode the context.
Seq2seq and HRED do not use background knowledge. 
\textbf{S2SA} is a model using attention mechanism to attend to background knowledge \citep{bahdanau2014neural}. 
\textbf{GTTP} leverages background information with a copying mechanism to copy a token from the background at the appropriate decoding step \citep{see2017get}.
\textbf{BiDAF} extracts a span from the background as response and uses a co-attention architecture to improve the span finding accuracy \citep{seo2016bidirectional}.

We use BLEU, ROUGE-1, ROUGE-2 and ROUGE-L as the automatic evaluation metrics.
Because the background knowledge and the corresponding conversations are restricted to a specific topic, there automatic evaluations are relatively more reliable for \acp{BBC} than for open-domain conversational modeling \citep{dinan2018wizard}. 
\section{Results and Analysis}

\subsection{Overall Performance}
We list the result of all methods in Table~\ref{tab:result}.

First, \ac{CaKe} outperforms the generation-based models, including \ac{S2S}, \ac{HRED}, \ac{S2SA} and \ac{GTTP}. 
Especially, for the strong baseline model \ac{GTTP}, \ac{CaKe} outperforms it by more than 35\% for all the metrics with regard to the oracle background. 
Meanwhile, \ac{CaKe} outperforms it by more than 20\% for each metric when it comes to the 256 words background. 
The improvements show that \ac{CaKe} is much better at locating the most relevant information in the background. 
The original \ac{GTTP} model uses the current decoder state to select knowledge, while \ac{CaKe} uses b2c and c2b attention to do pre-selection of knowledge. 
The comparison of \ac{CaKe} and \ac{GTTP} suggests that knowledge pre-selection is superior to classic single-state attention which uses the current state to attend to knowledge from the background.

Second, \ac{CaKe} is superior to the \ac{BiDAF} model with oracle background for all the metrics for more than 8\% except BLEU and ROUGE-2 with multi-references. 
For the 256 words background, our model beats \ac{BiDAF} on ROUGE-1 and ROUGE-l metrics. 
This suggests that \ac{CaKe} can generate better responses than \ac{BiDAF}.
The main reason is that besides extracting knowledge from background like \ac{BiDAF}, \ac{CaKe} can also generate tokens from vocabulary to enhance response fluency.

Third, the performance of \ac{CaKe} reduces slightly when the background becomes longer, but the reduction is acceptable considering that for the 256 words background \ac{CaKe} is still slightly superior to \ac{BiDAF}.

\begin{table*}[!htb]
  \caption{Case studies.}
  \label{tab:case}
  \centering
  \begin{tabular}{lp{14.5cm}}
    \toprule
    \multirow{8}{1.5cm}{Background} & The Mist, what? A bit like The Fog, then. Stephen King's The Mist, oh, that makes it even worse. Directed by Frank Darabont, since when did he direct horror films? Okay, so he scripted Nightmare on Elm Street 3 and The Blob, not bad films, but not classics in any sense. Starring Thomas Jane, has anyone seen The Punisher. And, to cap it all, The Mist died a time. Love this movie, ooof that ending. Sometimes I feel like the only person who prefers the book ending. It's more expansive and leaves something to the imagination. Classic Horror in a Post Modern age. The ending was one of the best I've seen. `The Mist' is worth watching! My favorite character was Melissa Mcbride. My favorite character was the main protagonist, David Drayton.  \\
    \cmidrule{2-2}
     \multirow{3}{1.5cm}{Context} 
      & Speaker 1: Which is your favorite character in this?\\
      & Speaker 2: My favorite character was the main protagonist, David Drayton.\\
      & Speaker 1: What about that ending?\\
    \cmidrule{2-2}
    \multirow{3}{0cm}{Response} 
    & BiDAF: Classic horror in a post modern age.\\
    
    & GTTP: They this how the mob mentality and religion turn people into monsters.\\
    & \ac{CaKe}: One of the best horror films I've seen in a long, long time.\\

\midrule    
    
    \multirow{8}{1.5cm}{Background} & In many sites, even in IMDb, there are some pops-up screens that irritate me. What I do not understand is some comments of persons that should never watch this type of movie. What do they expect to see when they go to the movie theaters or buy/rent a DVD of `Scary Movie 3', directed by David Zucker and with Leslie Nielsen in the cast? An art movie, with hidden messages, an epic, a classic or a film with politically correct jokes? Honestly, if I did not like this genre, in which Michael Jackson is disguised as an alien. My favorite character was Brenda. I think it was hilarious. Do you remember who came out of box office \textdollar 110,000,082 awards BMI Film \& TV Awards 2004 James L. Venable MTV Movie + TV Awards 2004 Best Cameo\\
    
    \cmidrule{2-2}
    
     \multirow{3}{1.5cm}{Context} & Speaker 1: And this again proved brilliance.\\
      & Speaker 2: I totally loved this one. Tho the ending kinda weird but overall it gives me a creep. \\
      & Speaker 1: Do you have any idea, how much it made on box office?
      \\
    \cmidrule{2-2}
      
    \multirow{3}{1.5cm}{Response} 
    & BiDAF: \textdollar 110,000,082.\\
    & GTTP: It made \textdollar 110,000,082.\\
    & \ac{CaKe}: I think it grossed \textdollar 110,000,082.\\
    \bottomrule
  \end{tabular}
\end{table*}

\subsection{Knowledge Selection Visualization}
As shown in Figure~\ref{fig:attention}, we visualize the attention weights to highlight the differences between the results of \ac{GTTP} and \ac{CaKe}. 
The responses of GTTP and \ac{CaKe} are: \emph{`It was so bad that it's good'} and \emph{`I agree, Fun, August, action movie'} respectively. 
The result of \ac{CaKe} is closest to the ground truth. 

The results of context-to-background attention show that attention is very strong on several positions of the background including the 92-th position, which is the token `fun' and suggests that the utterance history could help find the most relevant positions of the background. 
It suggests that our pre-selection mechanism could help knowledge selection. 
This explains why our model could select relevant knowledge from the background better than \ac{GTTP}.

\subsection{Case Study}
We select two examples from the test dataset to illustrate our results, as shown in Table~\ref{tab:case}. 
The examples suggest that \ac{CaKe} is able to generate more fluent responses than \ac{BiDAF} and more informative responses than \ac{GTTP}. 
First, \ac{CaKe} generates responses that are more relevant than \ac{GTTP}. 
Second, the responses of \ac{CaKe} are more natural and fluent than BiDAF, which can be inferred from the responses of BiDAF including \emph{`Classic horror in a post modern age'} and \emph{`\textdollar110,000,082'}. 
The reason is that \ac{BiDAF} extracts spans from the background sentences as responses directly.

There are also occasions that \ac{CaKe} does not perform well.
For instance, \ac{CaKe} generates common tokens like \emph{`I agree'} and \emph{`I know'} very frequently. 
This suggests that the diversity of the model needs to be taking into consideration. 


\section{Conclusion and Future Work}
In this paper, we propose knowledge pre-selection process for the \ac{BBC} task.
The proposed model, \ac{CaKe}, explores selecting relevant knowledge by using context as prior query. 
Experiments show that \ac{CaKe} outperforms the state-of-art method.
  
A limitation of \ac{CaKe} is that the performance of our pre-selection process decreases when the background becomes longer.
To further improve \ac{CaKe} in knowledge selection, we will explore alternative approaches to improve the selector and generator module in future work, such as multi-agent learning, transformer models and other attention mechanisms. 
As another future work, although automatic evaluations are relatively reliable for \acp{BBC}, it would be better if we can also conduct human evaluations.
Meanwhile, we also hope to improve the diversity of \ac{CaKe} by incorporating mechanisms such as changing optimization objects and leveraging mutual information~\citep{jiang-2019-improving}.

\smallskip\noindent%
\textbf{Acknowledgments.}
This research was partially supported by 
Ahold Delhaize,
the Association of Universities in the Netherlands (VSNU),
the China Scholarship Council (CSC),
and
the Innovation Center for Artificial Intelligence (ICAI).
All content represents the opinion of the authors, which is not necessarily shared or endorsed by their respective employers and/or sponsors.

\section*{Code}
To facilitate reproducibility of the results in this paper, we are sharing the code at \url{https://github.com/repozhang/bbc-pre-selection}.

\bibliographystyle{named}
\bibliography{scai2019-yangjun}

\end{document}